\documentclass{article}

\usepackage{graphicx} 
\usepackage{subfigure} 

\usepackage{natbib}

\usepackage{algorithm}
\usepackage{algorithmic}

\usepackage{amssymb}
\usepackage{amsmath}
\usepackage{bm}
\usepackage{mathtools}
\usepackage{paralist}
\usepackage{bbm}
\usepackage{dsfont}
\usepackage{units}

\graphicspath{ {figures/} }

\usepackage{hyperref}

\usepackage[accepted]{icml2013}
\usepackage{pgf}

\newenvironment{claim}[1]{\par\noindent\underline{Claim:}\space#1}{}

\begin{document} 

\twocolumn[
\icmltitle{Consistent Position Bias Estimation without Online Interventions for Learning-to-Rank}

\icmlauthor{Aman Agarwal}{aa2398@cornell.edu}
\icmladdress{Cornell University}
\icmlauthor{Ivan Zaitsev}{iz44@cornell.edu}
\icmladdress{Cornell University}
\icmlauthor{Thorsten Joachims}{tj@cs.cornell.edu}
\icmladdress{Cornell University}

\icmlkeywords{Unbiased learning to rank, counterfactual inference, propensity estimation}

\vskip 0.3in
]

\begin{abstract} 
Presentation bias is one of the key challenges when learning from implicit feedback in search engines, as it confounds the relevance signal with uninformative signals due to position in the ranking, saliency, and other presentation factors. While it was recently shown how counterfactual learning-to-rank (LTR) approaches \cite{Joachims/etal/17a} can provably overcome presentation bias if observation propensities are known, it remains to show how to accurately estimate these propensities. In this paper, we propose the first method for producing consistent propensity estimates without manual relevance judgments, disruptive interventions, or restrictive relevance modeling assumptions. We merely require that we have implicit feedback data from multiple different ranking functions. Furthermore, we argue that our estimation technique applies to an extended class of Contextual Position-Based Propensity Models, where propensities not only depend on position but also on observable features of the query and document. Initial simulation studies confirm that the approach is scalable, accurate, and robust.
\end{abstract} 

\section{Introduction}
In most information retrieval (IR) applications (e.g., personal search, scholarly search, product search), implicit user feedback (e.g. clicks, dwell time, purchases) is routinely logged and constitutes an abundant source of training data for learning-to-rank (LTR). However, implicit feedback suffers from presentation biases, which can make its naive use as training data highly misleading \cite{Joachims/etal/07a}. For example, the position at which a result is displayed introduces a strong bias, since higher-ranked results are more likely to be discovered by the user than lower-ranked ones.

It was recently shown that counterfactual inference methods provide a provably unbiased and consistent approach to LTR despite biased data \cite{Joachims/etal/17a}. The key prerequisite for counterfactual LTR is knowledge of the propensity of obtaining a particular feedback signal, which enables unbiased empirical risk minimization (ERM) via inverse propensity scoring. This makes getting accurate propensity estimates a crucial bottleneck for effective LTR, which is the problem we address in this paper.

In this work, we propose the first method for producing consistent propensity estimates without manual relevance judgments, disruptive interventions, or restrictive relevance modeling assumptions. We focus on propensity estimation under the Position-Based Propensity Model (PBM), but also consider a new and richer class of Contextual PBM (CPBM). In the Contextual PBM, examination of a result does not only depend on its rank, but can also depend on side information describing the query (e.g. navigational vs. informational) and the document under consideration (e.g. number of bolded snippet terms). The key idea behind our estimation technique is to exploit data from a natural intervention that is readily available in virtually any operational system -- namely that we have implicit feedback data from more than one ranking function. Since click behavior depends jointly on examination and relevance, we show how to exploit this intervention to control for any difference in overall relevance of results at different positions under both the PBM and the CPBM model. Crucially, we find that the rankers need not be drastically different in their result placements or overall performance, and we propose a technique to recover the relative propensities globally even if most of the changes in rank are small.

Our approach overcomes the problems of existing propensity estimation methods. First, conventional estimation approaches for the PBM as a generative click model \cite{Chuklin/etal/15a} require that individual queries repeat many times, which is unrealistic for many ranking setting. Second, to avoid this requirement, \cite{Wang/etal/18} include a relevance model. Unfortunately, they found that this leads to biased propensity estimates in practice, since defining an accurate relevance model is at least as hard as the learning-to-rank problem itself. Third, the gold standard standard for propensity estimation so far has been an intervention where the result in rank 1 is randomly swapped to any rank k \cite{Joachims/etal/17a}. While this provides provably consistent propensity estimates for the PBM, it degrades retrieval performance and user experience. The approach presented in this paper overcomes this disadvantage by leveraging existing data without a need for additional online interventions, while preserving statistical consistency and extending the expressiveness of the model to the CPBM.



\section{Setup}

We model user queries as sampled i.i.d. $q \sim \Pr(\mathrm{Q})$. Whenever a query is sampled, the ranker sorts the (pre-determined) candidate results $d$ for the query and displays the ranking to the user. Suppose query $q$ is sampled and result $d$ is displayed at position $k$. Let $C$, $E$ and $R$ be random variables corresponding to user behavior events of clicking, examination and relevance judgment for query-document pair $(q,d)$. Then according to the Position-Based Propensity Model (PBM) \cite{Chuklin/etal/15a},
\begin{align*}
    P(C=1|q,d,k) = P(E=1|k) P(R=1|q,d).
\end{align*}
In this model, the examination probability $p_k := P(E=1|k)$ depends only on the position, and it is identical to the observation propensity \cite{Joachims/etal/17a}. For learning, it is sufficient to estimate relative propensities $\nicefrac{p_k}{p_1}$ for each $k$ \cite{Joachims/etal/17a}, which is the goal in this paper. 

We also introduce the following Contextual Position-Based Model (CPBM), in which the examination probabilities at ranks 2 and beyond additionally depend on a context $x$, i.e.  
\begin{align*}
    \forall k \!\ge\! 2: P(\!C\!=\!1|q,d,k,x\!) = P(\!E\!=\!1|k,x\!) P(\!R\!=\!1|q,d\!).
\end{align*}
In this model, $x$ can include observable side information about the query $q$ and document $d$, and the following estimation method can be extended to the CPBM. However, we stick to the vanilla PBM for the sake of simplicity.

Now, suppose $m$ rankers $f_i$ were used in the past. A mild but crucial condition is that the choice of ranker $f_i$ must not depend on the query, which is analogous to exploration scavenging \cite{Scavenging2008}. Each ranker $f_i$ generated a click log $\mathcal{D}_i =\{q_i^j, y_i^j, \delta_i^j\}$ of size $n_i$.  Here $j\in [n_i]$, $q_i^j$ is a sampled query, $y_i^j$ the presented ranking and $\delta_i^j$ the vector of user feedback on each document in the ranking.  Furthermore, we denote $C_i^j$ as the candidate set of results for query $q_i^j$, $\mathrm{rk}(d|y_i^j)$  as the position or rank of candidate result $d$ in ranking $y_i^j$, and $\delta_i^j(d) \in [0,1]$ for whether result $d$ was clicked or not. 

\section{Method}

We begin by defining ``interventional'' sets of (query-document) pairs. Specifically, for each $k \neq k' \in [M]$ where $M$ is some fixed number of top positions for which estimates are desired (e.g. $M=10)$, let 
\begin{align*}
    S_{k,k'} := \{(q,d) : \exists f,\!f' \;  \mathrm{rk}(d|f\!(q)) \!=\! k \wedge \mathrm{rk}(d|f'\!(q)) \!=\! k'\}
\end{align*}
Intuitively, the pairs in these sets are informative because they receive different treatments or interventions by the different rankers. Next, define a weighting function $w(q,d,k)$ for each query, document and position in the following way
\begin{align*}
    w(q,d,k) := \sum_{i=1}^m n_i\mathbbm{1}[\mathrm{rk}(d|f_{i}(q)) = k]
\end{align*}
and define the following quantity for each $k \neq k' \in [M]$ 
\begin{align*}
    \hat{c}_k^{k,k'} \!\!:=\!\! \sum_{i=1}^m \!\sum_{j=1}^{n_i} \!\sum_{d \in C_i^j} \!\!\mathbbm{1}_{[(q_i^j,d) \in S_{k,k'}]}\mathbbm{1}_{[\mathrm{rk}(d|y_i^j)=k]}\frac{\delta_i^j(d)}{w(q_i^j,d,k)}.
\end{align*}
Note that $w(q_i^j,d,k)$ is non-zero whenever the first indicator is true. Then we make the following claim:
\begin{claim}
Denoting the expectation over queries $q \sim \Pr(\mathrm{Q})$ sampled i.i.d and user feedback $\delta$ according to the Position-Based Model (drawn for each sample in the logs) as  $\mathbb{E}_{q,\delta}[\cdot]$, for each $k \neq k' \in [M]$ 
\begin{align*}
    \frac{\mathbbm{E}_{q,\delta}[\hat{c}_k^{k,k'}]}{\mathbbm{E}_{q,\delta}[\hat{c}_{k'}^{k,k'}]} = \frac{p_k}{p_k'}.
\end{align*}
\end{claim}  

The proof involves writing out the expected values and taking terms common appropriately so that everything other than the propensities cancels out in the ratio, and we omit it here. 

Informally, $\hat{c}_k^{k,k'}$ captures the weighted click-through rate at position $k$ restricted to ($k$,$k'$)-interventional (query, document) pairs, where the weights $w(q,d,k)$ account for the disbalance in applying the intervention of putting document $d$ at position $k$ vs $k'$ for query $q$. Furthermore, in expectation, $\hat{c}_k^{k,k'}$ equals $p_k$ times $r_{k,k'}$, where $r_{k,k'}$ is the expected value of $P(R=1|q,d)$ restricted to the interventional set $S_{k,k'}$. Similarly, in expectation $\hat{c}_{k'}^{k,k'}$ equals $p_{k'}$ times $r_{k,k'}$, giving us the claim. Intuitively, we have controlled for relevance by restricting to interventional pairs. 

Note that simply using $\nicefrac{\hat{c}_k^{k,1}}{\hat{c}_1^{k,1}}$ gives consistent estimates of the target relative propensities $\nicefrac{p_k}{p_1}$, but this fails to use all (if not most) of the logged data due to the restriction to the $S_{k,1}$ interventional set. So, we propose an MLE based approach to tackle this issue. To do so, we first define the following ``no-click'' counterpart of $\hat{c}_k^{k,k'}$, 
\begin{align*}
    \hat{\neg c}_k^{k,k'} := \sum_{i=1}^m \sum_{j=1}^{n_i} \sum_{d \in C_i^j} \mathbbm{1}_{[(q_i^j,d) \in S_{k,k'}]}\mathbbm{1}_{[\mathrm{rk}(d|y_i^j)=k]}\frac{1-\delta_i^j(d)}{w(q_i^j,d,k)}.
\end{align*}
Finally, the following claim gives us our method
\begin{claim}
Let $\hat{p}_k$ and $\hat{r}_{k,k'}$ for $k\neq k' \in [M]$ be parameters that maximize the following objective
\begin{align*}
\sum_{k\neq k' \in [M]} \hat{c}_k^{k,k'} \log(\hat{p}_k \hat{r}_{k,k'}) + \hat{\neg c}_k^{k,k'} \log(1-\hat{p}_k \hat{r}_{k,k'}).
\end{align*}
Then $\nicefrac{\hat{p}_k}{\hat{p}_1}$ is a consistent estimate of relative propensity $\nicefrac{p_k}{p_1}$ for $k\in [M]$.
\end{claim}  

\begin{figure}
    \centering
    \includegraphics*[width=0.91\linewidth,trim={0.5cm 0cm 0.4cm 0.2cm},clip]{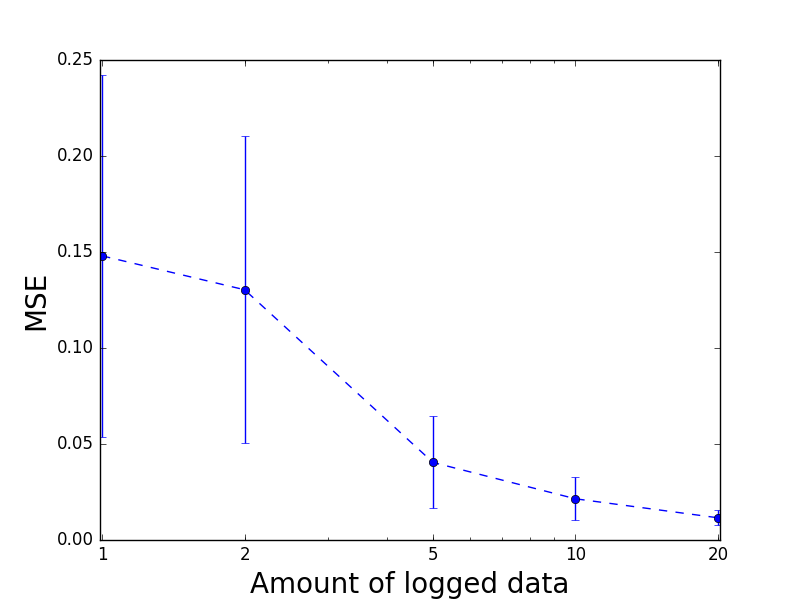}
    \vspace*{-0.3cm}
    \caption{Estimation error with increasing number of sweeps of the dataset during click simulation. ($\eta=1$, $\epsilon_-=0.1$, $\mathrm{overlap}=0.8$)}
    \label{fig:plot_n}
\end{figure}

The proof is omitted. Informally, we have that the expected values of $\hat{c}_k^{k,k'}$ and $\hat{\neg c}_k^{k,k'}$ equal $p_kr_{k,k'}$ and $1-p_kr_{k,k'}$ respectively ($r_{k,k'}$ as described earlier). So, the objective can be interpreted as an MLE problem which is consistent. 

Note that this approach uses the weighted click-through rates for every interventional pair, and further, a particular (query-document) pair may contribute to multiple interventional sets. Thus, the available data is being fully utilized. 

\section{Empirical Evaluation}

\begin{figure}
    \centering
    \includegraphics*[width=0.91\linewidth,trim={0.5cm 0cm 0.4cm 0.2cm},clip]{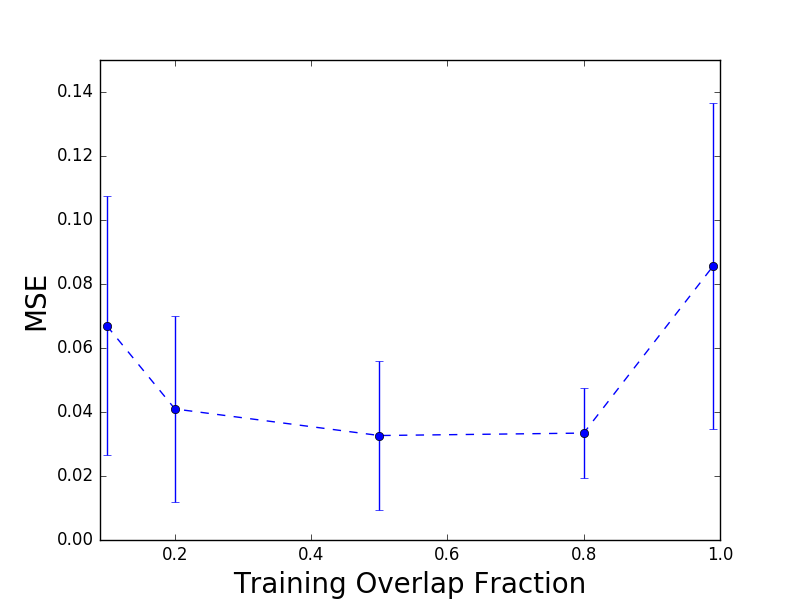}
    \vspace*{-0.3cm}
    \caption{Estimation error with increasing fraction of overlap in the training data for the rankers. ($\eta=1$, $\epsilon_-=0.1$, $\mathrm{sweeps}=5$)}
    \label{fig:plot_ov}
\end{figure}

We conducted various synthetic experiments on the Yahoo LTR Challenge corpus to demonstrate the accuracy, scalability and robustness of our method. 

Rankers were obtained by training Ranking SVMs on random slices of the full-information training set. The click logs were generated by simulating the Position-Based Model with propensities that decay with the presented rank of the result as $p_r = \big ( \frac{1}{r} \big )^\eta$. The parameter $\eta$ controls the severity of bias, with higher values causing greater position bias. We also introduced noise into the clicks by allowing some irrelevant documents to be clicked. Specifically, an irrelevant document ranked at position $r$ by the production ranker is clicked with probability $p_r$ times $\epsilon_-$ whereas a relevant document is clicked with probability $p_r$. For simplicity (and without loss of generality), we used click logs from two rankers in each experiment setting. The ``similarity'' of the two rankers was controlled by varying the degree of overlap in their respective training slices. We report the mean squared error in estimating the vector of relative propensities that were used in the click simulation upto rank $10$. Error bars indicate the variance over $6$ independent runs. We keep the slices used for training the rankers at 2\% of the full dataset. 

Figure~\ref{fig:plot_n} shows that the estimation accuracy consistently improves as the amount of logged data from the two rankers increases, supporting our theoretical claim of statistical consistency. In Figure~\ref{fig:plot_ov}, we see that the estimation accuracy remains quite robust even as the rankers become increasingly similar due to the overlap in the data they are trained on. As expected, the error goes up when the rankers are very similar since then they tend to put documents at the same position, leading to fewer interventional pairs. Interestingly, the error is also relatively higher when the rankers are too dissimilar. This is because when the candidate sets are larger than $10$, the dissimilarity in the rankers causes many interventions to be discarded since they often go beyond rank $10$. Finally, note that the estimation is robust to our chosen noise model. In fact, since we do not make any assumptions about result relevance, our method is applicable for any complex relevance plus noise model as long as the PBM holds.  

\section{Acknowledgments}
This work was supported by NSF awards IIS-1615706 and IIS-1513692, and through a gift from Amazon. This material is based upon work supported by the National Science Foundation Graduate Research Fellowship Program under Grant No. DGE-1650441. Any opinions, findings, and conclusions or recommendations expressed in this material are those of the author(s) and do not necessarily reflect the views of the National Science Foundation.

\bibliography{ref}
\bibliographystyle{icml2013}

\end{document}